\documentclass[journal,transmag]{IEEEtran}

\usepackage{amsmath}
\usepackage{indentfirst}
\usepackage{chngpage}
\usepackage{algorithmic}
\usepackage{wrapfig}
\usepackage{caption} 
\usepackage{mathtools}

\usepackage{cite}
\usepackage{lipsum}

\usepackage[ruled,vlined]{algorithm2e}
\usepackage{graphicx}
\usepackage[utf8]{inputenc} 
\usepackage[T1]{fontenc}    
\usepackage{hyperref}       
\usepackage{url}            
\usepackage{booktabs}       
\usepackage{amsfonts}       
\usepackage{nicefrac}       
\usepackage{microtype}      
\usepackage{bbding}         
\usepackage{hyperref}
\usepackage{xcolor}
\usepackage{tikz}
\usetikzlibrary{tikzmark}
\definecolor{lime}{HTML}{A6CE39}
\DeclareRobustCommand{\orcidicon}{%
    \begin{tikzpicture}
    \draw[lime, fill=lime] (0,0) 
    circle [radius=0.16] 
    node[white] {{\fontfamily{qag}\selectfont \tiny ID}};
    \draw[white, fill=white] (-0.0625,0.095) 
    circle [radius=0.007];
    \end{tikzpicture}
    \hspace{-2mm}
}

\foreach \x in {A, ..., Z}{%
    \expandafter\xdef\csname orcid\x\endcsname{\noexpand\href{https://orcid.org/\csname orcidauthor\x\endcsname}{\noexpand\orcidicon}}
}
\newcommand{\orcid}[1]{\href{https://orcid.org/#1}{\textcolor[HTML]{A6CE39}{\orcidicon}}}
\newcommand*{\email}[1]{%
    \normalsize\href{mailto:#1}{#1}\par
    }

\begin{document}

\title{Q-SMASH: Q-Learning-based Self-Adaptation of \\Human-Centered Internet of Things}

\author{\IEEEauthorblockN{Hamed~Rahimi\IEEEauthorrefmark{1,2,3}\orcid{0000-0001-9179-8625},
Iago~Felipe~Trentin\IEEEauthorrefmark{1,3}\orcid{0000-0002-1558-5668},
Fano~Ramparany\IEEEauthorrefmark{1}, 
Olivier~Boissier\IEEEauthorrefmark{3}\orcid{0000-0002-2956-0533}}
\IEEEauthorblockA{\IEEEauthorrefmark{1}Orange Labs, Meylan, France}
\IEEEauthorblockA{\IEEEauthorrefmark{2}Univ. Lyon, Universite Jean Monnet, Saint-Etienne, France}
\IEEEauthorblockA{\IEEEauthorrefmark{3}Mines Saint-Etienne, Univ. Clermont Auvergne, CNRS, \\UMR 6158 LIMOS, Institut Henri Fayol, Saint-Etienne, France}

\thanks{\textbf{Corresponding Author:} \email{hamed.rahimi@orange.com}}

}

\IEEEtitleabstractindextext{%
\begin{abstract}
 As the number of Human-Centered Internet of Things (HCIoT) applications increases, the self-adaptation of its services and devices is becoming a fundamental requirement for addressing the uncertainties of the environment in decision-making processes. Self-adaptation of HCIoT aims to manage run-time changes in a dynamic environment and to adjust the functionality of IoT objects in order to achieve desired goals during execution. SMASH is a semantic-enabled multi-agent system for self-adaptation of HCIoT that autonomously adapts IoT objects to uncertainties of their environment. SMASH addresses the self-adaptation of IoT applications only according to the human values of users, while the behavior of users is not addressed. This article presents Q-SMASH: a multi-agent reinforcement learning-based approach for self-adaptation of IoT objects in human-centered environments. Q-SMASH aims to learn the behaviors of users along with respecting human values. The learning ability of Q-SMASH allows it to adapt itself to the behavioral change of users and make more accurate decisions in different states and situations.
\end{abstract}

}

\maketitle

\IEEEpeerreviewmaketitle

\section{Introduction}

\IEEEPARstart{T}{he} Human-Centered Internet of Things (HCIoT) \cite{hciot} has an enlarging scope of activities spanning from sensing, computing to acting and even more, learning, reasoning and planning. This scope of activities focuses on the accessibility of interactive IoT systems to human beings and is integrated with various aspects of social life. As the number of HCIoT applications increases, the self-adaptation of these objects is becoming a fundamental requirement for addressing the uncertainties of the environment in decision-making processes\cite{muccini2018self}. Self-adaptation of HCIoT aims to manage run-time uncertainties in a dynamic environment and to adjust the functionality of IoT objects in order to achieve desired goals during execution. SMASH \cite{smash} is a multi-agent approach for self-adaptation of HCIoT that autonomously adapts IoT objects to uncertainties of their environment. This process is based on the integration of value-reasoning and goal-reasoning mechanisms to planning and acting \cite{ghallab2004automated} of a multi-agent system \cite{wooldridge2009introduction}. The architecture of SMASH is based on a value-driven reasoning and deliberation process that consists of 4 layers: 1) Value-Reasoning Layer, 2) Goal-Reasoning Layer, 3) Planning Layer, and 4) Acting Layer. The first layer is dedicated to reasoning upon human values, ordering them according to the context of the environment and the user's personal preferences. The second is dedicated to goal reasoning and identifying goals to be achieved, based on the context and values from the first layer. The third layer is for run-time planning given the context of the environment,  selected goals, and the given values respectively from the second and first layers. And the last layer is for initiation of acting upon given plans and values, selecting actions to be performed. SMASH agents are user-centric and respect 19 human values defined by Theory of Basic Human Values \cite{schwartz1992universals}. This theory has been supported by recent cross-national tests \cite{schwartz2012overview}, and results show that human values can be organized in a continuum based on the compatibility of their pursuit. 
\begin{figure}
  \centering
  \includegraphics[scale=0.8]{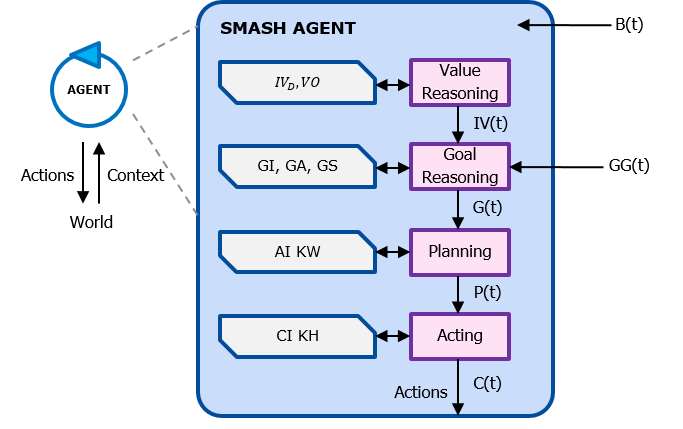}
  \caption{SMASH Agent Architecture Overview}
\end{figure}
Respecting human values is remarkably essential in designing the self-adaptation of HCIoT due to its close incorporation with humans and their environment. Although, we must keep in mind that people does not always behave aligned to their human values in their regular life, and this behavior of users is not necessarily a negative point in their personal life. For instance, according to the value base of some people, they are not obligated to accept a phone call from work, while they are on vacation or off duty. However, some of them accept phone calls most of the time due to the importance of their responsibilities. An intelligent system must be aware of such situations and be able to respond to the demand of users not only based on their values but also based on their behavior. In SMASH, we were able to address the self-adaptation of IoT applications only according to the human values of users. Therefore, SMASH needs a learning component that learns the behaviors of users along with respecting human values.

Considering the dynamics of HCIoT applications that address decision-making tasks, Reinforcement Learning is the best approach to perceive and interpret the environment, take actions and learn through trial and error. Reinforcement Learning (RL) \cite{rl} is a problem formulation for sequential decision making under uncertainty in which an agent produces its own information and learns the consequence of his own action through direct interaction within an environment. Reinforcement Learning  focuses on maximizing a reward signal without requiring exemplary supervision or complete models of the environment. 

In RL, the transition probability distribution defines what state the agent will reach on the next time step and the reward function defines how much reward the agent will receive on that time step. Considering that SMASH handles Planning and Acting, and its information only describes a part of the environment, we must use an RL technique that does not use the transition probability distribution and the reward function. Q-learning \cite{q} is an off-policy Temporal Difference (TD) \cite{td} control algorithm that is considered as a model-free reinforcement learning technique. Q-Learning investigates state-action pairs and learns the value of an action in a particular state. In Q-Learning, we create a Q-Table in which rows are possible states and columns are possible actions. The values in the Q-Table is calculated through the following formula and Algorithm 1.
{\small
\[
\underbrace{\text{New }Q(s,a)}_{\substack{\text{New}\\
                                          \text{Q-Value}}}
    = \underbrace{Q(s,a)}_{\substack{\text{Current}\\
                                     \text{Q-Value}}}
    + \tikzmarknode{A}{\alpha}
    \Bigl[
        \underbrace{R(s,a)}_{\text{Reward}}
    + \tikzmarknode{B}{\gamma}
        \overbrace{\max Q'(s',a')}^{\mathclap{%
            \substack{\text{Maximum predicted reward, given} \\
                      \text{new state and all possible actions}}
                                            }}
    - Q(s,a)
    \Bigr]
\begin{tikzpicture}[overlay, remember picture,shorten <=1mm, font=\footnotesize, align=center]
\draw (A.south) -- ++ (0,-.8) node (C) [below] {Learning\\ rate};
\draw (B.south) -- (B |- C.north)  node[below] {Discount\\ rate};
\end{tikzpicture}
\vspace{4ex}
\]
}

In this paper, we feed all the plans of SMASH to Q-Learning aiming to create a model of the environment that respects their human values and at the same time learns the behavior of users corresponding to those values. Q-SMASH is able to address various uncertainties of the HCIoT environment that includes human values and human behavior. The rest of the paper is organized as follows. In Section II, we will propose the structure of the Learning Layer in Q-SMASH. In Section III, we will discuss the results and the differences from the related works, and finally, we will conclude the paper in section IV. 

\begin{algorithm}[b!]
\small
Algorithm parameters: step size $\alpha \in (0, 1]$, small $epsilon > 0$\;
Initialize $Q(s, a)$, for all $s \in \mathcal{S}^+, a \in \mathcal{A}(s)$, \\arbitrarily except that $Q(\mathrm{terminal}, \cdot) = 0$\;
\ForEach{episode}{
    Initialize S\;
    \ForEach{step of episode}{
        Choose $A$ from $S$ using policy derived from $Q$\;
        Take action $A$, observe $R$, $S'$\;
        $Q(S, A) \leftarrow Q(S, A) + \alpha [R + \gamma \max_a Q(S', a) - Q(S, A)]$\;
        $S \leftarrow S'$\;
    }
}
\caption{Q-learning (off-policy TD control)}
\end{algorithm}

\section{Main Propose}
SMASH creates the world, the dynamics of the environment, and performs planning and acting for goal-driven IoT based on human values and the preference of users. Although SMASH can address the uncertainty of the environment according to human values, it is not capable of considering the behavior of users. In this approach, which is called Q-SMASH, we integrate Q-Learning to SMASH and create a model of the environment that respects human values and at the same time learns the behavior of users corresponding to those values. This integration decreases the number of trial and error for exploration of different state-action pairs for Q-Learning and improves the efficiency of SMASH by learning the behavior of users thanks to Q-Learning.

\begin{figure}[t!]
  \centering
  \includegraphics[width=1\linewidth]{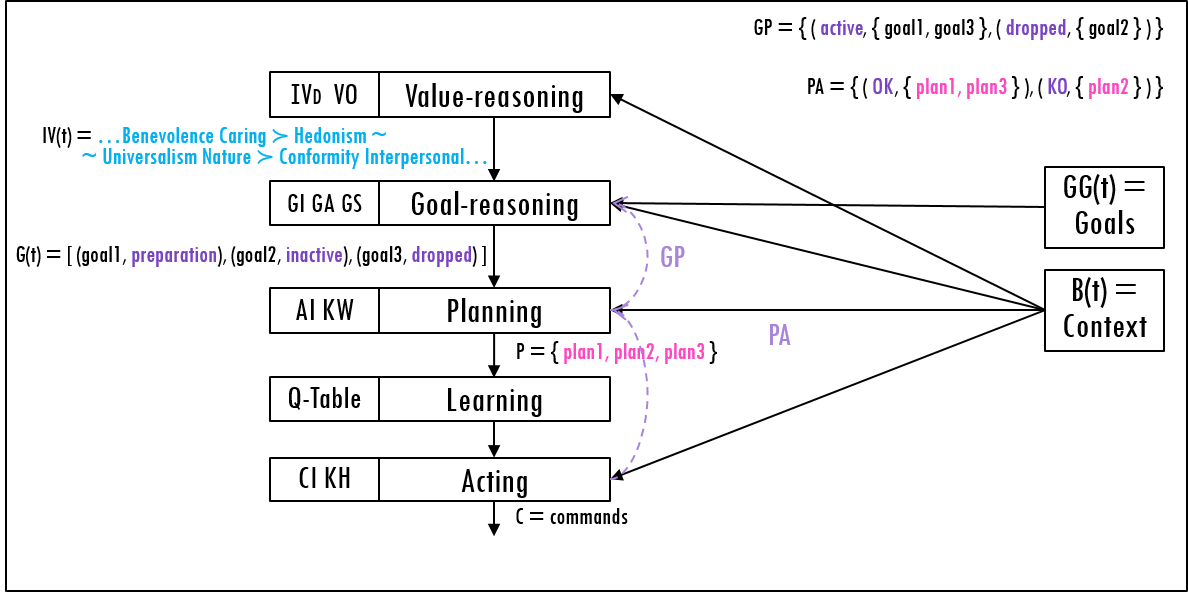}
  \caption{Q-SMASH Agent Architecture}
\end{figure}

As you can see in Fig 2, the architecture of Q-MASH has one layer more comparing to SMASH. This layer is called Learning Layer and is between the Planning layer and Acting layer. This additional layer is responsible to both learn the behavior of users and respect human values. The Learning layer is consisting of four components as shown in Fig 3: 1) Decision Maker, 2) State-Action Prediction, 3) Reward Function, and 4) Q-Update. All the components are explained as follows.

\begin{figure}
  \centering
  \includegraphics[scale=0.6]{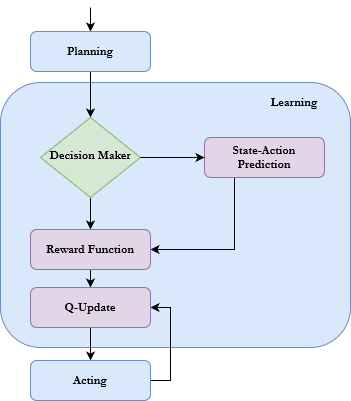}
  \caption{Learning Layer Overview}
\end{figure}

\subsection{Decision Maker}
The Planning provides the Learning layer with a set of plans that satisfies the user's goals. These plans are consisting of the current and next states of devices that are based on the belief base of the user. The Decision Maker component is responsible to decide if the actions will be based on plans provided by the Planning Layer or to predict actions based on the current state of devices. Plenty of methods can be used for Decision Maker. It can be a neural network or a fuzzy algorithm. In this layer, we use a dynamic epsilon-greedy algorithm that generally makes decisions based on plans provided by the Planning Layer at the beginning and then converges by time to predicting the actions based on the current state of devices. As shown in Algorithm 2, the Epsilon-Greedy algorithm is an approach to balance exploration and exploitation by choosing between exploration and exploitation randomly. The dynamic-epsilon-greedy refers to the epsilon greedy algorithm that has a dynamic epsilon, which gets updated after an episode by a set of parameter such as time or number of steps. In our design, we consider a scenario as a episode and we train the learning layer based on the repeating of the scenario. We update the amount of epsilon after each episode.

\begin{algorithm}
\small
\hspace*{\algorithmicindent} \textbf{Input} next\_state-action, current\_state \\
Initialize State-Action, Epsilon = $0<\epsilon<1$\;
\ForEach{episode}{
    \ForEach{step of episode}{
    P = Random(0,1)\;
    \eIf{$(P<\epsilon)$}{
        State-Action=St-Action\_Predic(current\_state)\;
        }
    { 
    State-Action=next\_state-action\;
        }
        Reward\_Function(State-Action)\;
    }
    Update\_Epsilon()\;
}
\caption{Dynamic Epsilon-Greedy Algorithm for Decision Maker}
\end{algorithm}

\subsection{State-Action Prediction}
This component is responsible to predict action corresponding to the current state. There are several approaches to perform the prediction. we can diverse these approaches into two types: 1) Collaborative and 2) Non-collaborative. In collaborative approaches, the agent has a direct interaction with humans and receives straightforward feedbacks from the user. For instance, this kind of interaction can be performed by interactive audio systems such as Orange Djingo, Amazon Alexa, or Google Home. In non-collaborative approaches, there are three ways to predict the action without the interference of the user corresponding to the current state. The first is to randomly choose an action and investigate its impact on the environment and the preference of the user. The second way is to perform policy evaluation from Q-Learning and find the best action based on the behavior of the user. And finally, the third way is to use advanced algorithms based on deep learning and predict the action. In our design, we use a collaborative approach for the sake of simplicity and directly receive the best action from the user's feedback through the Phone. Finally, we pass the chosen action to the Reward Function.  

\subsection{Reward Function}

As shown in Algorithm 3, the Reward Function is responsible to compute the reward corresponding to the action chosen by State-Action Prediction or directly received by Planning Layer. If action is chosen by Planning Layer, then Reward Function will add a small positive reward to the system in order to learn the SMASH. Indeed, with this positive reward, we continually learn a part of the environment that is based on the values of users. If we receive the action from State-Action Prediction, then we will compare it with the action received from Planning Layer. In case the two actions are the same thing, we will add a big reward to the system to understand the behavior of users in that state is completely aligned with their values. However, in the case the action from State-Action Prediction and the action received by Planning Layer are not the same, we will understand that the user prefers to act differently than his values in that state. Considering that we have assumed the system aims to adapt itself to the behavior the system will have two Q-update: First, we add a positive reward for the action chosen by State-Action Prediction to adapt the system based on the behavior. Second, we will mention a negative reward for the action received from the Planning Layer to decrease the impact of the values of users on the self-adaptation of the system.

\begin{algorithm}
\small
\hspace*{\algorithmicindent} \textbf{Input} Action\_Planning, Action\_SAP, Action\_Source, current\_state\;

    \eIf{$Action\_Source == State-Action\_Prediction$}{
        \eIf{$Action\_Planning == Action\_SAP$}{
        Update\_Q(current\_state,Action\_SAP, R=+5)\;
        }
        {
        Update\_Q(current\_state, Action\_Planning, R=-5)\;
        Update\_Q(current\_state, Action\_SAP, R=+5)\;
        }
        }
        {
        Update\_Q(current\_state, Action\_Planning, R=+1)\;
        }
\caption{Reward Function Algorithm}
\end{algorithm}

\subsection{Q-update}

The Reward Function provides this component with Current State, Action, and Reward. First, Q-Update computes the Next State using the Action and the Current State. Second, It will encode the actions and the state of devices, which have string data types, to position values in the Q-table, which have integer data types. Finally, it updates the new Q-values by feeding the encoded Current State, encoded Next State, encoded Action, and Reward to Q-Value Function and using the mathematics explained in the previous section. Q-update is provided in Algorithm 4.
\begin{algorithm}[h!]
\small
\textbf{Input} Current\_State, Next\_State, Reward\;

        Action= Action\_model(Current\_State, Next\_State);  \\
        encoder(Current\_State, Next\_State, Action) \\
        $Q(S, A) \leftarrow Q(S, A) + \alpha [R + \gamma \max_a Q(S', a) - Q(S, A)]$\;
        $S \leftarrow S'$\;

\caption{Q-Update Algorithm}
\end{algorithm}

\section{Q-SMASH and IoT Layer}
Smart Home applications are increasingly becoming larger and more complex due to the spread of various ubiquitous appliances that are supported by different vendors. In addition to the increase in the number of devices, the stack of messages, data formats, communication protocols, and system architectures are even so evolving. Facing such heterogeneity, several standardizations such as ETSI SmartM2M SAREF \cite{lefranccois2017planned} and W3C Web of Things (WoT) \cite{kovatsch_web_2019} try to address the syntactic and semantic languages of these systems. Existing solutions to integrate different IoT services are mainly based on manual IoT mashups (e.g. NodeRed) \cite{guinard2009towards}, which are a way to compose a new service from existing ones \cite{kovatsch2015practical} in order to allow the synergy among heterogeneous devices, services, protocols, and formats. Even in ad-hoc mashups that are often temporary, heterogeneity is limited, and services are provided for specific needs \cite{guinard2009towards}. Some solutions discover devices and their associated services using existing Web infrastructure. Although, these solutions face several issues such as connection and integration. Semantic Web Technologies (SWT) \cite{berners2001semantic} are a high-level contextual language that abstracts communications and creates a common request format for accessing different services on the web infrastructure. Semantic-enabled IoT systems are an effective solution that addresses the complexity of context management \cite{euzenat2008dynamic,perera2013context} and improves interoperability among heterogeneous devices. This interoperability allows the system to have access to various contextual information, which helps to take advantage of various adaptation techniques that require more contextual resources. 

This section presents the integration between Q-SMASH and the Home’In platform, which is able to interact with various devices in Smart Homes and address their interoperability issues. Fig. 4 shows the proposed approach which is consisting of a multi-agent system and Home’In platform that is connected to real devices. The approach firstly represents a multi-agent system composed of Q-SMASH autonomous agents that are responsible to satisfy users’ goals considering their values and behavior in real-time. Following a multi-agent oriented programming approach \cite{boissier2013multi}, the agents have uniform access to the resources and tools of the environment via a set of non-autonomous entities called artifacts. In Smart Home applications, artifacts are responsible to encapsulate the access and control of smart devices such as TV, Phone, or Sofa.  In Home’in, the Context Manager receives information from the physical world via devices connected to Home’In. These devices frequently send data to dedicated Home’In service providers that are responsible to consider different abstraction layers with various formats and content, and then store them in the Context Manager as RDF triples to update the system in real-time. In order to provide context for the agent, we created an interface with an already existing industrial solution working as a context manager. This industrial solution provides two channels for retrieval and storage of contextual information:
\begin{itemize}
\item  a REST API with an HTTP endpoint for context retrieval, that receives SPARQL requests and replies with JSON or XML responses (both using predefined ontologies selected by the industry responsible for the solution); and
\item an MQTT broker for context storage, that receives and replies JSON messages coded in proprietary format.
\end{itemize}


Our contextual information is accessed through request/response (e.g. REST) or publish/subscribe (e.g. MQTT) patterns. The IoT Devices Manager provides a direct interface to contact the devices that are located in the physical environment of the home and are connected to Home’In. In other words,  it sends commands to these devices in order to make them perform actions or to retrieve the information they’ve sensed in the environment. Home’In components are interconnected through MQTT and Rest buses that enable the Context Manager to listen to abstraction layers data and store selected ones through IoT Devices Manager for representation of the context.

\begin{figure}[t]
\centering
\includegraphics[width=.85\linewidth]{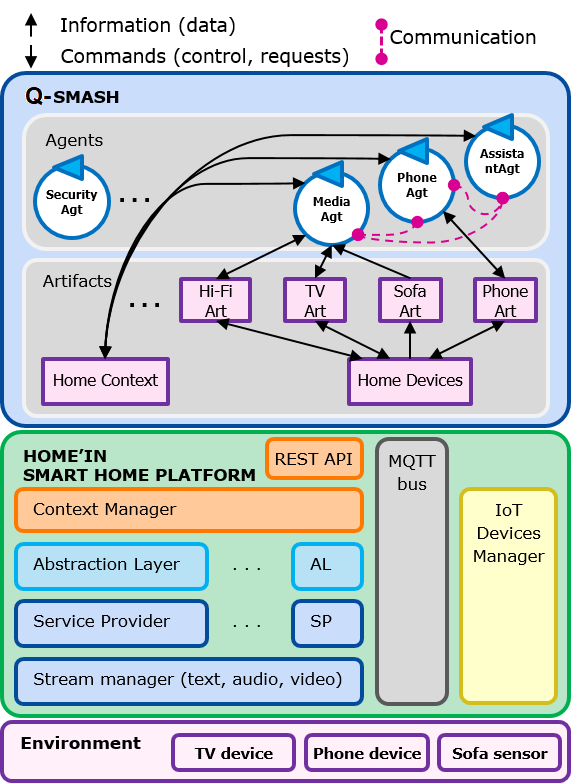}
\caption{Integration of Q-SMASH and Home'In}
\end{figure}

\section{Related works and Discussion}
The main objective of the paper is to propose an additional layer to SMASH in order to put forward the research interest and development of smart environments, especially the intelligent home, a residential space that takes into accounts more subjective user aspects such as human values and human behavior, creating an intelligent autonomous home that better understands its users and makes more accurate decisions. There are several works that focus on the AI of Smart Homes using Reinforcement Learning. In \cite{set}, the authors propose to integrate Cognitive Dynamic Systems (CDS) with Smart Homes based on Reinforcement Learning and Bayesian Filter. The authors of \cite{situ} use Q-Learning to build a task-recommendation service for daily activities in Smart Homes. In \cite{jarvis}, the authors use Deep Reinforcement Learning to optimize temperature, used energy, and cost. The main difference between the aforementioned works and Q-SMASH is that the human values are respected via the value-reasoning process of Q-SMASH. The next possible step is adding a DQN (Deep Q-Network) agent that has learning capability via Deep Reinforcement Learning to the proposed multi-agent system. This agent is capable of observing the behavior of the user more precisely.

\section{Conclusion}
In this paper, we proposed one additional layer to SMASH, which is a semantic-enabled approach for self-adaptation of IoT objects in human-centered smart environments. This layer is based on Q-Learning and is able to autonomously adapt the functionality of IoT devices According to the behaviour of users. In SMASH, we were able to address the self-adaptation of IoT applications only according to the human values of users. Q-SMASH is able to address various uncertainties of the HCIoT environment that includes human values and human behavior and make more accurate decisions in different states and situations.

\section*{Acknowledgment}

This  research  is  supported  by  the  grant  of Orange Labs, Meylan. We would like to thank Vincent Rouault and Prof. Pierre Maret for providing their support and  cooperation  throughout  the  research.

\ifCLASSOPTIONcaptionsoff
  \newpage
\fi


\begin{thebibliography}{0}

\bibitem{smash}
Rahimi, Hamed, et al. "SMASH: a Semantic-enabled Multi-agent Approach for Self-adaptation of Human-centered IoT." accepted in PAAMS'21, Available on arXiv:2105.14915 (2021).

\bibitem{rl}
Sutton, Richard S., and Andrew G. Barto. Reinforcement learning: An introduction. MIT press, 2018.

\bibitem{jarvis}
Mudgerikar, Anand, and Elisa Bertino. "Jarvis: Moving Towards a Smarter Internet of Things." 2020 IEEE 40th International Conference on Distributed Computing Systems (ICDCS). IEEE, 2020.

\bibitem{situ}
Oyeleke, Richard O., Chen-Yeou Yu, and Carl K. Chang. "Situ-centric reinforcement learning for recommendation of tasks in activities of daily living in smart homes." 2018 IEEE 42nd Annual Computer Software and Applications Conference (COMPSAC). Vol. 2. IEEE, 2018.

\bibitem{set}
Feng, Shuo, Peyman Setoodeh, and Simon Haykin. "Smart home: Cognitive interactive people-centric Internet of Things." IEEE Communications Magazine 55.2 (2017): 34-39.
\bibitem{hciot}
Meshram, Chandrashekhar, et al. "A Lightweight Provably Secure Digital Short-Signature Technique Using Extended Chaotic Maps for Human-Centered IoT Systems." IEEE Systems Journal (2020).

\bibitem{muccini2018self}
Muccini, Henry, et al. "Self-adaptive IoT architectures: An emergency handling case study." Proceedings of the 12th European Conference on Software Architecture: Companion Proceedings. 2018.

\bibitem{ghallab2004automated}
Ghallab, Malik, Dana Nau, and Paolo Traverso. Automated Planning: theory and practice. Elsevier, 2004.

\bibitem{wooldridge2009introduction}
Wooldridge, Michael. An introduction to multiagent systems. John wiley \& sons, 2009.

\bibitem{schwartz2012overview}
Schwartz, Shalom H. "An overview of the Schwartz theory of basic values." Online readings in Psychology and Culture 2.1 (2012): 2307-0919

\bibitem{schwartz1992universals}
Schwartz, Shalom H. "Universals in the content and structure of values: Theoretical advances and empirical tests in 20 countries." Advances in experimental social psychology. Vol. 25. Academic Press, 1992. 1-65.
\bibitem{q}
Watkins, C. J., \& Dayan, P. (1992). Q-learning. Machine learning, 8(3-4), 279-292

\bibitem{td}
Tesauro, G. (1992). Practical issues in temporal difference learning. Machine learning, 8(3), 257-277.

\bibitem{lefranccois2017planned}
Lefrançois, Maxime. "Planned etsi saref extensions based on the w3c\&ogc sosa/ssn-compatible seas ontology paaerns." Workshop on Semantic Interoperability and Standardization in the IoT, SIS-IoT. 2017.

\bibitem{kovatsch_web_2019}
Kovatsch, R. Matsukura, M. Lagally, T. Kawaguchi, K. Toumura, and K. Kajimoto. WEBOF THINGS AT W3C, May 2019.

\bibitem{guinard2009towards}
Guinard, Dominique, et al. "Towards physical mashups in the web of things." 2009 Sixth International Conference on Networked Sensing Systems (INSS). IEEE, 2009.

\bibitem{kovatsch2015practical}
Kovatsch, Matthias, Yassin N. Hassan, and Simon Mayer. "Practical semantics for the Internet of Things: Physical states, device mashups, and open questions." 2015 5th International Conference on the Internet of Things (IOT). IEEE, 2015.

\bibitem{euzenat2008dynamic}
Euzenat, Jérôme, Jérôme Pierson, and Fano Ramparany. "Dynamic context management for pervasive applications." Knowledge engineering review 23.1 (2008): 21-49.

\bibitem{perera2013context}
Perera, Charith, et al. "Context aware computing for the internet of things: A survey." IEEE communications surveys \& tutorials 16.1 (2013): 414-454.

\bibitem{berners2001semantic}
Berners-Lee, Tim, James Hendler, and Ora Lassila. "The semantic web." Scientific american 284.5 (2001): 34-43.

\bibitem{boissier2013multi}
Boissier, Olivier, et al. "Multi-agent oriented programming with JaCaMo." Science of Computer Programming 78.6 (2013): 747-761.


\end{thebibliography}
\end{document}